\pdfoutput=1

\documentclass[11pt]{article}

\usepackage[preprint]{acl}

\usepackage{times}
\usepackage{latexsym}

\usepackage[T1]{fontenc}

\usepackage[utf8]{inputenc}

\usepackage{microtype}

\usepackage{inconsolata}

\usepackage{graphicx}

\title{The \#Somos600M Project: Generating NLP resources that represent the diversity of the languages from LATAM, the Caribbean, and Spain}

\author{María Grandury \\
    SomosNLP \\
    \texttt{mariagrandury@somosnlp.org} \\}

\begin{document}

\maketitle
\begin{abstract}
We are 600 million Spanish speakers. We launched the \#Somos600M Project because the diversity of the languages from LATAM, the Caribbean and Spain needs to be represented in Artificial Intelligence (AI) systems.
Despite being the 7.5\% of the world population, there is no open dataset
to instruction-tune large language models (LLMs), nor a leaderboard to evaluate and compare them.
In this paper, we present how we have created as an international open-source community the first versions of the instruction and evaluation datasets,  indispensable resources for the advancement of Natural Language Processing (NLP) in our languages.
\end{abstract}

\section{Introduction}

We are
600 million Spanish speakers\footnote{Combining natives
and foreign language learners.} according to the 2023 Yearbook of the Cervantes Institute \citep{anuario-instituto-cervantes}.
However, we lack native instruction-tuned LLMs -- models fine-tuned on a collection of NLP tasks formatted as instructions. This fine-tuning improves their zero-shot capabilities \citep{weifinetuned} and adaptability, relevant for AI alignment, chat-like interactions and retrieval augmented generation (RAG) applications.
The \#Somos600M Project\footnote{somosnlp.org/somos600m}, led by SomosNLP\footnote{SomosNLP.org is a community of Spanish speakers whose mission is to achieve fair representation of Spanish and co-official languages in the digital world.}, aims to create the necessary resources to fine-tune and evaluate these language models.

Spanish is the official language in 22 countries, which implies the existence of a great number of geographic varieties or dialects and influences the performance of language models
\citep{bogantes-etal-2016-towards, castillo-lopez-etal-2023-analyzing}.

Moreover,
in these countries there are co-official languages from completely different language families like Quechua in Latin America (LATAM) and Euskera in Spain.
The scarcity of resources in these languages hinders the development of NLP applications \citep{hedderich}, which worsens the socioeconomic situation of these communities and the risk of extinction of some of these languages \citep{mager-etal-2018-challenges}.

With the \#Somos600M Project, we emphasize the importance of representing this diversity in LLMs. The initial objectives of the project are:

\begin{itemize}
    \vspace{-0.2cm}
    \item \textbf{Creation of an open instruction dataset:} A set of input-output pairs that represent various tasks, include different Spanish varieties and co-official languages, and enable the fine-tuning of LLMs to follow instructions.
    \vspace{-0.2cm}
    \item \textbf{Establishment of an open leaderboard:} Standardizing the evaluation of generative LLMs in Spanish and co-official languages by creating an open impartial leaderboard.
    \vspace{-0.2cm}
\end{itemize}

All the generated resources are open-source.\footnote{ huggingface.co/somosnlp}

\section{Previous Work}

Since the release of the first Spanish pre-trained language model, BETO \citep{CaneteCFP2020},
there has been a notable increment in the availability of open Spanish resources 
(Annex~\ref{sec:appendix-hf-hub})
thanks to various initiatives.\footnote{hf.co/spaces/somosnlp/spanish-nlp-initiatives}
There are also 
shared tasks
and projects that aim to create resources in Spanish dialects \citep{guevara-rukoz-etal-2020-crowdsourcing, hernandez-mena-meza-ruiz-2022-creating}, 
indigenous languages of LATAM
\citep{pendas-etal-2023-neural, 
ebrahimi-etal-2023-findings}, 
and co-official languages of Spain \citep{etxaniz2024latxa, gonzalez-aguirre-etal-2024-infraestructure}.

However, when it comes to instructions there is a void. Since 2020, we have seen a trend to fine-tune language models using English natural language instructions \citep{flan-collection} and there are more than 2,000 "instruct" datasets in the Hugging Face Hub\footnote{huggingface.co}.
To the best of our knowledge, in total in our languages there are 227k instructions in Catalan created by the AINA\footnote{projecteaina.cat} and ILENIA\footnote{proyectoilenia.es} projects and only 14k originally created in Spanish, 10k by these same projects and 4k corresponding to the AYA dataset by CohereForAI\citep{singh2024aya}.
This forces the Spanish-speaking community to use error-prone machine-translated English datasets directly or to manually validate them\footnote{hf.co/datasets/somosnlp/somos-clean-alpaca-es}.

We 
proposed the creation of instructions as the task for the 2024 edition of the SomosNLP Hackathon\footnote{somosnlp.org/hackathon}. 
The general goal of this recurring international online event is the creation of open-source NLP resources in Spanish and co-official languages, encouraging projects with societal impact related to the Sustainable Development Goals.
In 2022, we invited the community to fine-tune Transformer architecture models \citep{vaswani2017attention}, and in 2023, once again using LoRA-type techniques \citep{hu2022lora}, resulting in the publication of interesting projects \citep{Serrano2022BioMedIAAC, vasquez-rodriguez-etal-2022-benchmark}.

We also acknowledge the growing need for standard evaluation benchmarks in the field of Spanish language models. There is a new leaderboard, ODESIA\footnote{leaderboard.odesia.uned.es}, with 15 bilingual Spanish/English discriminative tasks, 
and another one, CLUB, for Catalan\footnote{club.aina.bsc.es}. Regarding text generation, we highlight recent assessments of LLMs' knowledge \citep{conde2024open, martínez2023words}. We propose the creation of an open leaderboard that evaluates different capabilities of generative models (e.g., domain knowledge, information extraction, linguistic proficiency, ethical aspects) in our languages and serves as a reference for the Spanish-speaking scientific community.

\section{The Project}

To create a large instruction dataset and a generative LLM leaderboard, we have launched several initiatives: an instruction generation hackathon, a dataset collection campaign, and an effort to translate and validate English evaluation datasets.

\subsection{Instruction generation}

During the SomosNLP 2024 Hackathon, the initial version of the large open instruction dataset was created. The participants had to generate synthetic instruction datasets for the later fine-tuning of an LLM with up to 7B parameters with QLoRA-like techniques \citep{dettmers2023qlora}.
Given the scarcity of resources across all topics, each team was free to choose the theme of their project. The hackathon was open to everyone (Annex~\ref{sec:appendix-hackathon}), regardless of prior NLP knowledge, and targeted individuals with both technical and linguistic backgrounds, encouraging interdisciplinary teams.

The teams had access to computing and storage resources, example notebooks, mentorship sessions, workshops, and talks\footnote{somosnlp.org/eventos} throughout March 2024 up to April 10th, in addition to visibility
and prizes to continue developing their NLP skills.

\subsection{Dataset collection}

In addition to generating new resources, reusing existing ones is crucial. Hence, we launched a dataset collection campaign, with a focus on Spanish dialects and co-official languages. Training datasets will be transformed into question-answer pairs \citep{keskar2019unifying}, while evaluation datasets will be included in the generative LLM leaderboard.

\subsection{Translation validation}

The Open LLM Leaderboard \citep{open-llm-leaderboard} stands as one of the most popular English LLM leaderboards, and some of its constituent datasets were machine-translated as part of the Okapi project \citep{dac2023okapi}. In collaboration with Hugging Face and Argilla, we launched a community effort for native Spanish speakers to validate these translations.
We also joined the international initiative Data Is Better Together (DIBT)\footnote{github.com/huggingface/data-is-better-together} to validate the translation of 500 prompts, in order to include Spanish in the corresponding future multilingual leaderboard.

\section{Results}

We present the results with respect to both objectives of the \#Somos600M Project.

\subsection{Instruction datasets}

18 projects were presented to the hackathon, resulting in a total of 2,333,052 examples created, summing up to 324 MB of data (Annex~\ref{sec:appendix-corpus-it}).

\begin{figure}[!ht]
  \includegraphics[width=\columnwidth]{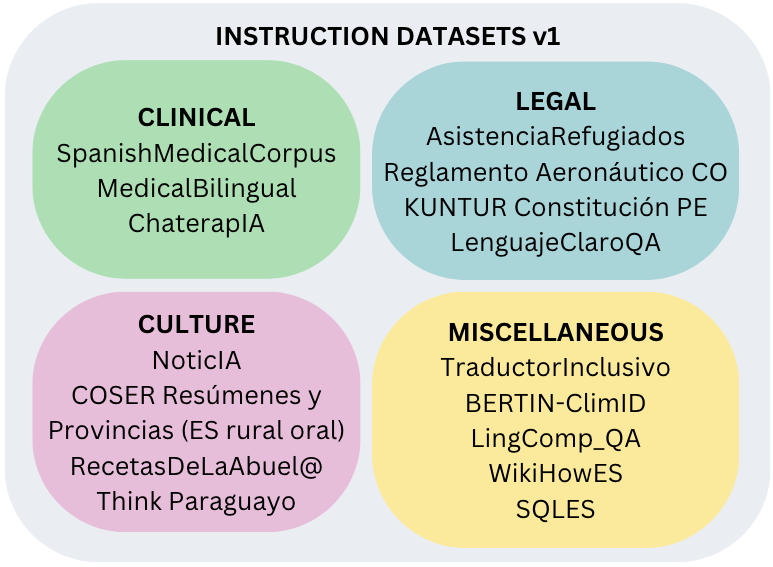}
  \caption{Instruction datasets generated during the \#Somos600M Hackathon grouped by domain.}
  \label{fig:egci}
\end{figure}

We highlight the high number of countries represented in the chosen topics (e.g., Colombian Aeronautical Regulation, Refugee Legal Assistance, Peruvian Constitution, international traditional recipes), as well as the project on Guarani culture.
Most teams focused on text models, except for one delving into the various accents of rural Spain.
A significant amount of data was generated in the healthcare and legal sectors 
(Figure~\ref{fig:egci}).
We also note projects involving clear and inclusive language rewriting, clickbait news summarization, and sustainability text detection.

\subsection{Evaluation datasets}

\begin{figure}[!ht]
  \includegraphics[width=\columnwidth]{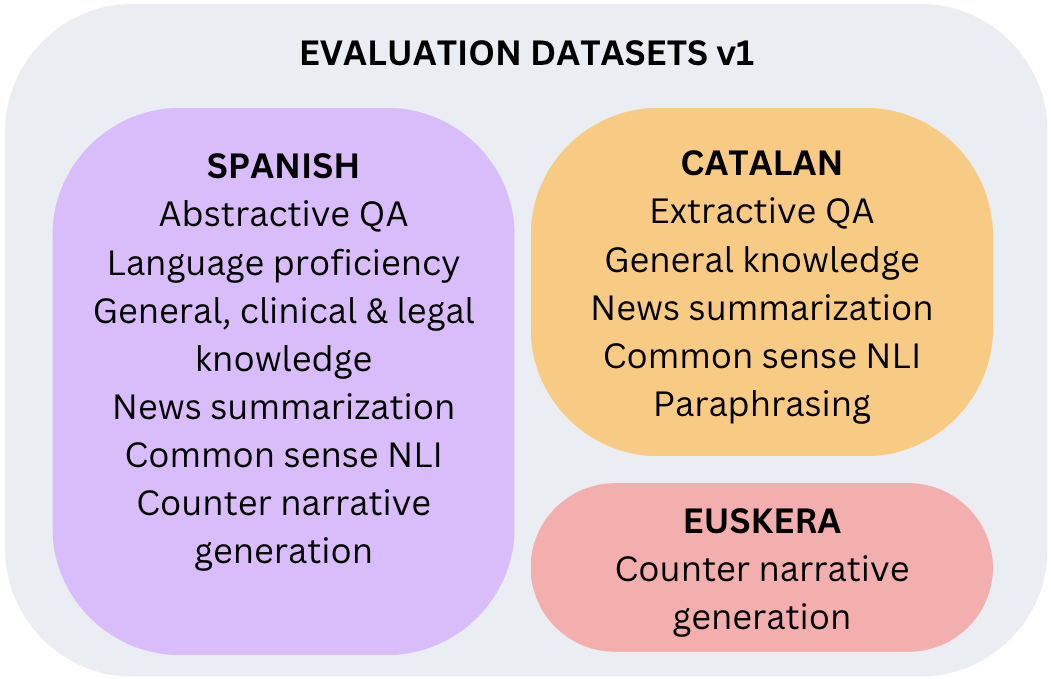}
  \caption{Tasks and languages (ES: Spanish, CA: Catalan and EU: Euskera) of the evaluation datasets of the first version of the open generative LLM leaderboard.}
  \label{fig:corpus-eval}
\end{figure}

In the first collection campaing round, we received the donation of 5 evaluation datasets manually annotated by experts and, in the second one, 14 more, adding new languages (Figure~\ref{fig:corpus-eval}). Together with the translations, they form the first version of the open generative LLM leaderboard (Annex~\ref{sec:appendix-corpus-eval}).

In the validation of the Okapi translations, a total of 61 persons participated, covering a 60\% of ARC-C \citep{Clark2018ThinkYH}, 15\% of HellaSwag \citep{zellers2019hellaswag} and 15\% of MMLU \citep{hendrycks2021measuring}.\footnote{hf.co/spaces/somosnlp/BenchmarkAnnotationDashboard}
Moreover, with the support of 37 persons, 100\% of DIBT's prompts were validated, which positioned Spanish as the first language to fully validate its translation.\footnote{hf.co/spaces/DIBT/PromptTranslationMultilingualDashboard}

\section{Discussion}

We are very proud of how the community answered our call. Generating 2 million instructions and gathering 22 evaluation datasets is a great advancement for instruction-tuned LLMs in our languages. 

Concerning the hackathon,
we are pleased to see that the number of datasets generated tripled last year's.

We confirm the usefulness of the libraries \texttt{distilabel}, \texttt{Argilla} and \texttt{transformers} to fine-tune LLMs with instructions synthetically created and manually reviewed. 
We highlight that a couple of teams also created their own annotation spaces
and asked the community for collaboration.

In the translation campaigns, we observed that most of the validation was conducted by 10\% of the individuals. For teams interested in organizing similar efforts (and for our next iteration), we recommend:
1) writing a clear annotation guide and enabling a feedback channel to iterate and improve the guide,
2) sharing an instructional video, and
3) creating a visualization of the initiative's progress to motivate and give visibility to the contributors.

\section{Conclusion}

The hackathon, the collection campaign, and the annotation efforts have enabled us to create the initial versions of the large instruction dataset and the open generative LLM leaderboard.

We are going to keep collaborating with entities from LATAM, the Caribbean, and Spain to organize hackathons focused on specific topics, varieties, and languages, scale up the collection campaign to create the most inclusive dataset possible, and expand the leaderboard by including evaluations of ethical (e.g., biases, hate speech) and linguistic (e.g., language variety adequacy) aspects, as well as other co-official languages.

The generated resources are open; we invite entities with greater computing power to use them for training (with our support, if desired) high-quality LLMs that are open, inclusive, and native.

\section*{Acknowledgments}
We thank all hackathon participants for their efforts. Thanks to their work, we now have the first version of a diverse instruction dataset. We thank Hugging Face for sponsoring the compute and storage resources, LenguajeNaturalAI, Cálamo \& Cran, and SaturdaysAI for providing prizes to motivate the participants, and LatinX in AI for inviting us to present our work to the LatinX in NLP workshop. Thank you also to the speakers for sharing their knowledge with the community.

With respect to the leaderboard, we thank Hugging Face and Argilla for co-organizing the translation validation efforts, and all the volunteers who participated in the annotation process. We are also thankful to the Instituto de Ingeniería del Conocimiento (IIC) de la Universidad Autónoma de Madrid (UAM), LenguajeNaturalAI, Grupo de Internet de Nueva Generación (GING) de la Universidad Politécnica de Madrid (UPM), Centro Vasco de Tecnología de la Lengua (HiTZ) and Barcelona Supercomputing Center (BSC) for donating high-quality evaluation datasets.

Thank you to everyone who helped review this paper, especially to Diana Galván, Flor Plaza, Abi Oppenheim, Pedro Reviriego, Javier Conde and Javier Aula-Blasco.

Finally, we extend our heartfelt gratitude to all those who generously volunteer their time to support our mission of democratizing NLP for the Spanish-speaking community.

\bibliography{anthology,custom}


\appendix

\section{Spanish resources in the Hugging Face Hub}
\label{sec:appendix-hf-hub}

\begin{figure}[!ht]
  \includegraphics[width=\columnwidth]{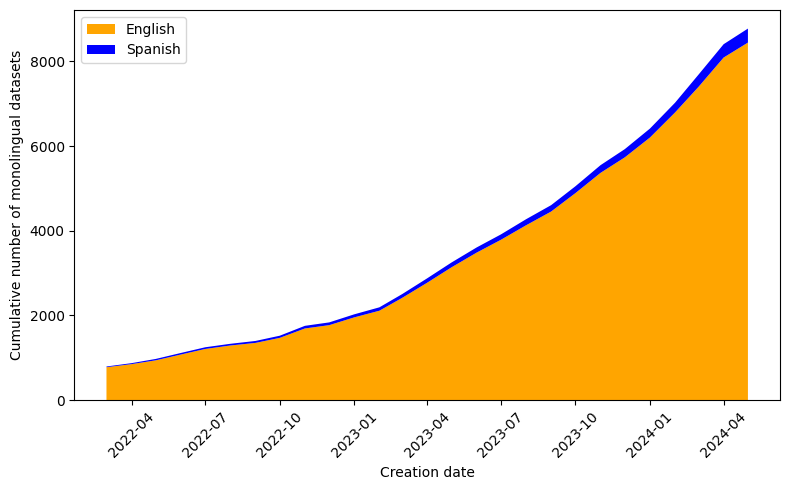}
  \caption{Cumulative number of monolingual English (orange) and Spanish (blue) datasets in the Hugging Face Hub over time until May 13 2024.}
  \label{fig:huggingface-hub-2}
\end{figure}

Even though the number of open-source NLP resources in Spanish on the Hugging Face Hub is increasing, the gap between Spanish and English remains substantial (Figure~\ref{fig:huggingface-hub-2}).


The code used to visualize this gap is available on the Hugging Face hub\footnote{hf.co/spaces/mariagrandury/language-gap-in-hf-hub}.

\section{Hackathon participants}
\label{sec:appendix-hackathon}

\begin{figure}[!ht]
    \includegraphics[width=\columnwidth]{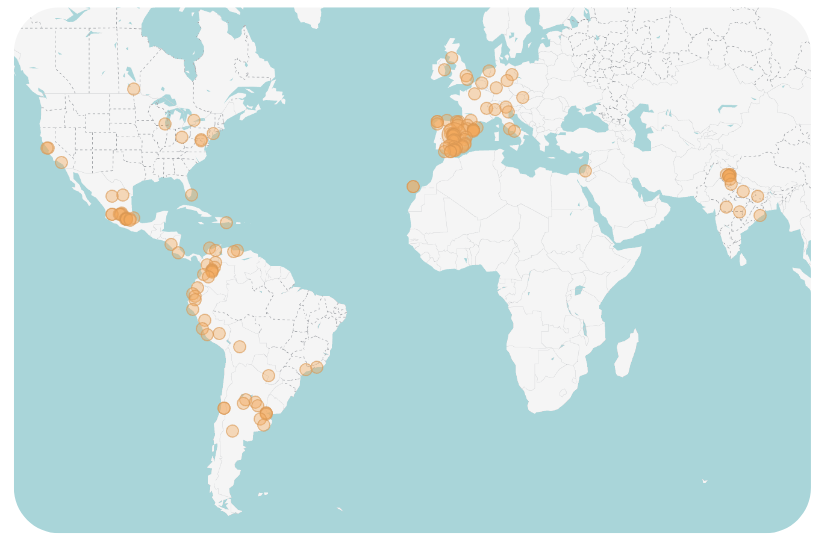}
  \caption{Location of the \#Somos600M Hackathon participants.}
  \label{fig:map}
\end{figure}

\begin{figure}[!ht]
    \includegraphics[width=\columnwidth]{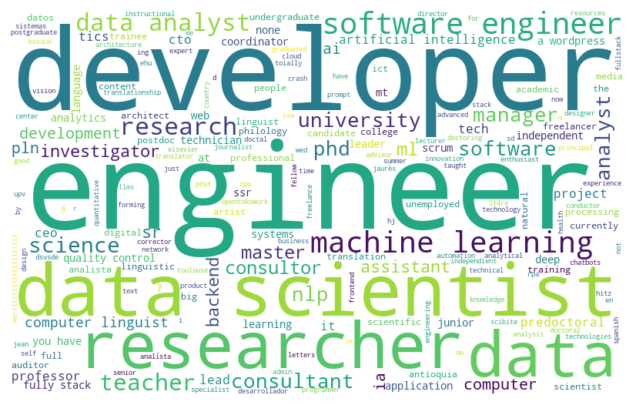}
  \caption{Wordcloud of the occupations of the Hackathon \#Somos600M participants translated to English.}
  \label{fig:wordcloud}
\end{figure}

651 people from 29 countries (Figure~\ref{fig:map}) registered for the hackathon, of which 222 were interested exclusively in attending the talks. It is worth noting that at least\footnote{The questions were optional.} 14\% of the individuals had already participated in one of the previous editions, almost 50\% of the participants were in LATAM (Table~\ref{tab:location}) and less than 40\% self-identify as non-male (Table~\ref{tab:gender}), these are numbers we would like to increase in future editions. With respect to the background and occupation of the participants, more than 40\% thought they had a fundamental NLP level before starting the hackathon (Table~\ref{tab:level}) and most of them are "developers", "engineers" or "data scientists" (Figure~\ref{fig:wordcloud}).

\begin{table}[!ht]
   \centering
  \begin{tabular}{ccc}
    \hline
    \textbf{ LATAM } & \textbf{ Spain } & \textbf{ Other }  \\
    \hline
    46\% & 32\% & 22\% \\
    \hline
  \end{tabular}
  \caption{Location of the \#Somos600M Hackathon participants, might not match the nationality.}
  \label{tab:location}
\end{table}

\begin{table}[!ht]
   \centering
  \begin{tabular}{cccc}
    \hline
    \textbf{ Female } & \textbf{ Male } & \textbf{ NB } & \textbf{ NR }  \\
    \hline
    22\% & 60\% & 1\% & 17\% \\
    \hline
  \end{tabular}
  \caption{Self-identified gender of the \#Somos600M Hackathon participants: "female", "male", "NB" (no binary) or "NR" (no response).}
  \label{tab:gender}
\end{table}

\begin{table}[!ht]
   \centering
  \begin{tabular}{cccc}
    \hline
    \textbf{ Fund. } & \textbf{ Intermediate } & \textbf{ Advanced } & \textbf{ NR }  \\
    \hline
    40\% & 31\% & 12\% & 17\% \\
    \hline
  \end{tabular}
  \caption{Self-assigned NLP level of the \#Somos600M Hackathon participants: "fundamental", "intermediate", "advanced", "NR" (no response).}
  \label{tab:level}
\end{table}

\section{Guidelines and resources}
\label{sec:appendix-hackathon-resources}

The guidelines of the hackathon\footnote{somosnlp.org/hackathon/bases} and the validation campaign\footnote{somosnlp.org/blog/validacion-benchmarks-argilla} can be found on the SomosNLP website. 

Since the hackathon was open to everyone and we expected participants from a wide range of levels, we organized mentorship sessions, hands-on workshops, and research talks. The recordings of all the sessions are available on YouTube\footnote{youtube.com/watch?v=JzpvHRrqtSU\&list=PLTA-KAy8nxaASMwEUWkkTfMaDxWBxn-8J} and the corresponding notebooks are available on GitHub\footnote{github.com/somosnlp/recursos/tree/main/hackathon\_2024}.

\section{Instruction dataset details}
\label{sec:appendix-corpus-it}

We enumerate the datasets created during the \#Somos600M Hackathon (Table~\ref{tab:instructions}):
\textbf{AsistenciaRefugiados}: Legal assistant for refugees or individuals in political asylum \citep{somosnlp2024asistenciarefugiados},
\textbf{BERTIN-ClimID}: BERTIN-Base Climate-related text Identification \citep{huerta2024climid},
\textbf{ChaterapIA}: Therapeutic conversations \citep{juliofc2024convterapeuticas},
\textbf{COSER Provincias}: Identification of provinces from oral interviews in rural Spain \citep{adsuar2024coserprovincias},
\textbf{COSER Resúmenes}: Summaries of oral interviews in rural Spain \citep{adsuar2024coserresumenes},
\textbf{KUNTUR}: Legal assistance in Peru \citep{quispe2024kuntur},
\textbf{LenguajeClaroQA}: Simplification of administrative language \citep{delafuente2024lenguajeclaro},
\textbf{LingComp\_QA}: Educational corpus about computational linguistics \citep{LingComp_QA},
\textbf{NoticIA}: Clickbait Article Summarization Dataset \citep{garcíaferrero2024noticia},
\textbf{RAC}: Colombian Aeronautical Regulation \citep{sepulveda2024rac},
\textbf{RecetasDeLaAbuel@}: Recipes from Spanish-speaking countries \citep{recetasdelaabuela2024},
\textbf{SMC}: Spanish Medical Corpus \citep{lopez2024spanishmedicallm},
\textbf{SQLES}: Interact with an SQL database in Spanish \citep{sql2024rangel},
\textbf{ThinkParaguayo}: Get to know the Guaraní culture \citep{paiva2024guarani},
\textbf{TraductorInclusivo}: Rewriting of texts using inclusive language \citep{GAMIJ2024es-inclusive-language},
and
\textbf{WikiHowES} \citep{quispe2024wikihowes}.

\begin{table*}
   \centering
  \begin{tabular}{lcccc}
    \hline
    \textbf{Dataset} & \textbf{\# Examples} & \textbf{MB} & \textbf{Domain} & \textbf{Country} \\
    \hline
    AsistenciaRefugiados & 10707 & 20.7 & Legal & ES, MX, VE + \\
    BERTIN-ClimID & 3680 & 1.63 & Sustainability & PE, ES \\
    ChaterapIA & 1000 & 2.30 & Psychology & ES \\
    COSER Provincias & 1150 & 0.22 & Rural culture & ES (oral) \\
    COSER Resúmenes & 230 & 1.08 & Rural culture & ES (oral) \\
    KUNTUR & 2075 & 0.73 & Legal & PE \\
    LenguajeClaroQA & 4094 & 1.72 & Legal admin. & ES \\
    LingComp\_QA & 1004 & 0.35 & Comp. Linguistics & ES \\
    MedicalBilingual & 8138 & 12.8 & Clinical & Mix \\
    NoticIA & 850 & 3.41 & Press & ES \\
    RAC & 24478 & 1.84 & Legal & CO \\
    RecetasDeLaAbuel@ & 20221 & 42.4 & Gastronomy & ES, MX, PE, AR+ \\
    SpanishMedicalCorpus & 2136490 & 48.5 & Clinical & ES, CL \\
    SQLES & 81 & 0.40 & Programming & - \\
    Think Paraguayo & 1498 & 0.19 & Guarani culture & PY \\
    TraductorInclusivo & 4196 & 0.40 & Miscellaneous & ES, AR, MX, CL, CR + \\
    WikiHowES & 113160 & 186 & Miscellaneous & Mix \\
    \hline
    Total & 2,333,052 & 324.67 & - & - \\
    \hline
  \end{tabular}
  \caption{Instruction datasets generated by the teams participating in the \#Somos600M Hackathon, available at huggingface.co/somosnlp. We excluded the dataset versions adapted to specific formats for model training (e.g. Gemma). The countries are represented by their corresponding code ISO 3166-1 alfa-2, the symbol "+" means that there are other countries represented in the dataset with a lower percentage.}
  \label{tab:instructions}
 
\end{table*}

\section{Evaluation dataset details}
\label{sec:appendix-corpus-eval}

The evaluation datasets included in the first version of the generative LLM leaderboard (Table~\ref{tab:corpus-eval}) are:
\textbf{AQuAS}: Abstractive Question-Answering in Spanish \citep{iic2024aquas} and \textbf{RagQuAS}: Retrieval-Augmented-Generation and Question-Answering in Spanish \citep{iic2024ragquas}, donated by the Instituto de Inge niería del Conocimiento (IIC) of the Universidad Autónoma de Madrid (UAM);
\textbf{SpaLawEx}: Spanish Law School Access Exams \citep{lenguajenaturalai2024spalawex}, \textbf{MedicalExpertES}: Diagnosis and treatment of clinical cases in Spanish \citep{lenguajenaturalai2024medicalexpertes} and \textbf{HumorQA}: White humour joke classification \citep{lenguajenaturalai2024humorqa}, donated by LenguajeNaturalAI;
\textbf{TELEIA}: Test de Español como Lengua Extranjera para Inteligencia Artificial, donated by the Next Generation Internet Group (GING) of the Universidad Politécnica de Madrid (UPM);
\textbf{Meta4XNLI}: A Crosslingual Parallel Corpus for Metaphor Detection and Interpretation \citep{sanchezbayona2024meta4xnli}, \textbf{NoticIA}: A Clickbait Article Summarization Dataset in Spanish \citep{garcíaferrero2024noticia}, \textbf{MedExpQA}: Multilingual Benchmarking of Large Language Models for Medical Question Answering
 \citep{alonso2024medexpqa}, \textbf{CasiMedicos-SQUAD}: Explanatory Argument Extraction of Correct Answers in Resident Medical Exams \citep{goenaga2023explanatory} and \textbf{CONAN-EUS}: Basque and Spanish Parallel Counter Narratives Dataset \citep{bengoetxea-et-al-2024} donated by the Basque Center for Language Technology (HiTZ); \textbf{CatalanQA}: Extractive QA in Catalan, \textbf{TE\_ca}: Textual Entailment in Catalan, \textbf{caBREU}: Article Summarization in Catalan, \textbf{XNLI\_ca}: Cross-lingual Natural Language Inference in Catalan, \textbf{COPA\_ca}: Choice Of Plausible Alternatives in Catalan, \textbf{PAWS\_ca}: Paraphrase Adversaries from Word Scrambling in Catalan, \textbf{XQUAD\_ca}: Cross-lingual Question Answering Dataset in Catalan and \textbf{WNLI\_ca}:  Winograd NLI dataset, created or professionally translated with funding of the projects AINA and ILENIA and donated by the Barcelona Supercomputing Center (BSC) \citep{gonzalez-aguirre-etal-2024-infraestructure}.

\begin{table*}[!ht]
  \centering
  \begin{tabular}{lccc}
    \hline
    \textbf{ Dataset } & \textbf{ Language } & \textbf{ Domain } & \textbf{ Task } \\
    \hline
    AQuAS & ES & Miscellaneous & Abstractive QA \\ 
    RagQuAS & ES & Miscellaneous & RAG and Abstractive QA \\ 
    HellaSwag\_es & ES & Miscellaneous & Commonsense NLI \\ 
    MMLU\_es & ES & Miscellaneous & Multiple choice QA \\
    TELEIA & ES & Language proficiency & Multiple choice QA \\
    Meta4XNLI & ES & Language proficiency & NLI \\ 
    HumorQA & ES & Language proficiency & Classification \\
    ARC-C\_es & ES & Science & Multiple choice QA \\
    NoticIA & ES & Press & Summarization \\
    SpaLawEx & ES & Legal & Multiple choice QA \\
    MedicalExpertES & ES & Clinical & Open QA \\
    MedExpQA & ES & Clinical & Multiple choice QA \\
    CasiMedicos-SQUAD & ES & Clinical & Extractive QA \\
    CONAN-EUS & ES, EU & Hate speech & Counter narrative generation \\
    CatalanQA & CA & Miscellaneous & Extractive QA \\ 
    TE\_ca & CA & Miscellaneous & NLI \\ 
    XNLI\_ca & CA & Miscellaneous & NLI \\ 
    WNLI\_ca & CA & Miscellaneous & NLI \\ 
    COPA\_ca & CA & Miscellaneous & Commonsense Reasoning \\ 
    PAWS\_ca & CA & Miscellaneous & Paraphrasing \\
    XQUAD\_ca & CA & Miscellaneous & Extractive QA \\ 
    caBREU & CA & Press & Summarization \\
    \hline
  \end{tabular}
  \caption{Datasets of the first version of the generative LLM leaderboard, that includes tasks in Spanish (ES), Catalan (CA) and Euskera (EU) and evaluates abstractive and extractive QA, general, clinical and legal knowledge, common sense reasoning, natural language inference (NLI) and language proficiency. 
  }
  \label{tab:corpus-eval}
\end{table*}

\newpage
\renewcommand{\thesection}{\arabic{section}}
\setcounter{section}{0}

\maketitleagain

\begin{abstract}
Somos 600 millones de hispanohablantes. Lanzamos el Proyecto \#Somos600M porque necesitamos que la riqueza de nuestras lenguas esté representada en los sistemas de Inteligencia Artificial (IA).
A pesar de ser el 7.5\% de la población mundial, no contamos con un gran corpus de instrucciones abierto que nos permita adaptar grandes modelos de lenguaje generativos, ni con una tabla de clasificación estándar para evaluarlos y compararlos. En este artículo presentamos cómo hemos creado en comunidad la primera versión de los corpus de instrucciones y de evaluación, recursos imprescindibles para el avance del Procesamiento de Lenguaje Natural (PLN) en nuestras lenguas.
\end{abstract}

\section{Introducción}

Somos cerca de 600 millones de hispanohablantes\footnote{Suma de los grupos de dominio nativo, competencia limitada y aprendices de lengua extranjera.} según el Anuario del Instituto Cervantes 2023 \citep{anuario-instituto-cervantes}.
Sin embargo, no contamos con grandes modelos de lenguaje (LLM, del inglés \textit{Large Language Model}) propios adaptados para seguir instrucciones (o \textit{prompts}). Esta adaptación mejora la versatilidad de los modelos \citep{weifinetuned}, importante para el alineamiento de la IA y aplicaciones de tipo conversacional y RAG (\textit{Retrieval Augmented Generation}).
El Proyecto \#Somos600M, liderado por SomosNLP\footnote{SomosNLP.org es una comunidad de hispanohablantes cuya misión es conseguir una justa representación del español y lenguas cooficiales en el mundo digital.}, tiene por objetivo crear los recursos necesarios para adaptar y evaluar estos modelos.

El español es lengua oficial en 22 países, lo que implica la existencia de una gran cantidad de variedades geográficas o dialectos, que influyen en el rendimiento de los modelos de PLN \citep{bogantes-etal-2016-towards, castillo-lopez-etal-2023-analyzing}.



Además, en estos países se hablan otras lenguas cooficiales de diferentes familias, como el quechua en LATAM y el euskera en España.
La escasez de recursos en estas lenguas dificulta el desarrollo de modelos de lenguaje \citep{hedderich}, lo que empeora la situación socioeconómica de estas comunidades y el riesgo de extinción de algunas de estas lenguas \citep{mager-etal-2018-challenges}.


Con el proyecto \#Somos600M hacemos hincapié en la representación de las variedades del español y las lenguas cooficiales en los recursos de PLN. Los objetivos iniciales del proyecto son:

\begin{itemize}
    \vspace{-0.2cm}
    \item \textbf{Crear un gran corpus abierto de instrucciones}: Un conjunto de pares pregunta-respuesta que represente las variedades del español y lenguas cooficiales y permita adaptar modelos que sigan instrucciones.
    \vspace{-0.2cm}
    \item \textbf{Crear una tabla de clasificación abierta}: Estandarizar la evaluación de modelos de lenguaje generativos en español y lenguas cooficiales mediante la creación de una tabla de clasificación abierta e imparcial.
    \vspace{-0.2cm}
\end{itemize}

Todos los recursos generados son abiertos.\footnote{huggingface.co/somosnlp}

\section{Antecedentes}



Desde la publicación del primer modelo de lenguaje pre-entrenado en español, BETO \citep{CaneteCFP2020},
hemos visto un aumento del número de recursos en español y lenguas cooficiales disponibles (Anexo~\ref{sec:appendix-hf-hub-es})
gracias a varias iniciativas.\footnote{hf.co/spaces/somosnlp/spanish-nlp-initiatives}. También hay talleres en congresos 
y proyectos para crear recursos en dialectos \citep{guevara-rukoz-etal-2020-crowdsourcing, hernandez-mena-meza-ruiz-2022-creating}, 
lenguas originarias de LATAM \citep{pendas-etal-2023-neural, ebrahimi-etal-2023-findings},
y lenguas cooficiales de España \citep{etxaniz2024latxa}.

En lo referente a instrucciones, desde 2020 hemos visto una tendencia a adaptar LLMs utilizando instrucciones en inglés \citep{flan-collection}
Sin embargo, hasta donde sabemos existen 227k en catalán creadas por AINA\footnote{projecteaina.cat} e ILENIA\footnote{proyectoilenia.es} y solo 14k originalmente creadas en español, 10k por estos proyectos y 4k de la iniciativa AYA\citep{singh2024aya}.
Esto obliga a la comunidad hispanohablante a utilizar instrucciones traducidas automáticamente del inglés o a validarlas manualmente\footnote{hf.co/datasets/somosnlp/somos-clean-alpaca-es}.

Así, propusimos la creación de instrucciones como tarea del Hackathon SomosNLP 2024\footnote{somosnlp.org/hackathon}. El objetivo general recurrente de este
evento internacional en línea es la creación de recursos abiertos de PLN en español y lenguas cooficiales, con enfoque en impulsar proyectos con impacto social. En 2022 hicimos una llamada a la comunidad para adaptar modelos con arquitectura Transformer \citep{vaswani2023attention} mediante fine-tuning y en 2023 con técnicas tipo LoRA \citep{hu2022lora}, resultando en la publicación de proyectos interesantes \citep{Serrano2022BioMedIAAC, vasquez-rodriguez-etal-2022-benchmark}.

Reforzamos nuestra contribución atendiendo a la necesidad creciente de evaluar nuestros modelos.
Existe una nueva tabla clasificación para modelos discriminativos, ODESIA\footnote{leaderboard.odesia.uned.es}, y una para catalán, CLUB\footnote{club.aina.bsc.es}.
Respecto a la generación de texto, destacamos recientes evaluaciones del conocimiento de los LLMs \citep{conde2024open, martínez2023words}.
Nuestra propuesta es crear una tabla de clasificación abierta que evalúe diferentes capacidades de los modelos generativos (e.g., dominio de un tema, extracción de información, adecuación lingüística, aspectos éticos) y sirva de referencia para la comunidad científica hispanohablante.


\section{El Proyecto}

Para crear un gran corpus de instrucciones y una tabla de clasificación de modelos generativos hemos lanzado varias iniciativas: un hackatón de generación instrucciones, una campaña de recolección de corpus y la traducción y validación de corpus de evaluación en inglés.

\subsection{Generación de instrucciones}

Aprovechamos el Hackathon SomosNLP 2024 para crear la primera versión del gran corpus abierto de instrucciones.
La tarea de los equipos era generar instrucciones sintéticas para la posterior adaptación de un LLM de hasta 7B con técnicas tipo QLoRA \citep{dettmers2023qlora}.
Dada la escasez general de recursos
dejamos a elección de cada equipo el tema concreto de su proyecto.
El hackatón estaba abierto a todo el mundo (Anexo~\ref{sec:appendix-hackathon-es}), sin requerimientos de conocimientos previos de PLN, y dirigido a personas tanto de formación técnica como lingüística, animando la creación de equipos interdisciplinares. 

Los equipos participantes tuvieron acceso durante el mes de marzo de 2024 y hasta el 10 de abril a recursos de computación y almacenamiento,
tutoriales, mentorías, talleres y charlas\footnote{somosnlp.org/eventos}, además de visibilidad
y premios para seguir formándose.

\subsection{Recolección de corpus}

Además de generar nuevos recursos, es importante reutilizar los existentes. Lanzamos una campaña de recolección de corpus\footnote{somosnlp.org/donatucorpus}, con especial foco en las diferentes variedades del español y lenguas cooficiales. Las bases de datos de entrenamiento se utilizarán para generar pares pregunta-respuesta \citep{keskar2019unifying}. Las de evaluación se incluirán en la tabla de clasificación de modelos generativos.

\subsection{Validación de traducciones}

La \textit{Open LLM Leaderboard} \citep{open-llm-leaderboard} es una de las tablas de clasificación más populares para evaluar LLMs en inglés y algunas de las bases de datos que la constituyen fueron traducidas automáticamente como parte del proyecto Okapi \citep{dac2023okapi}.
Lanzamos con Hugging Face y Argilla una iniciativa para validar en comunidad dichas traducciones.
También nos sumamos a la iniciativa internacional \textit{Data Is Better Together} (DIBT) \footnote{github.com/huggingface/data-is-better-together} para validar la traducción de 500 instrucciones y que el español forme parte de la correspondiente futura tabla de clasificación multilingüe.

\section{Resultados}

Exponemos los resultados respecto a los dos objetivos del Proyecto \#Somos600M.

\subsection{Corpus de instrucciones}

Se presentaron al hackatón 18 proyectos
y en total se crearon 2,333,052 ejemplos, que se traducen en 324 MB de datos (Anexo~\ref{sec:appendix-corpus-it-es}).

\begin{figure}[!ht]
  \includegraphics[width=\columnwidth]{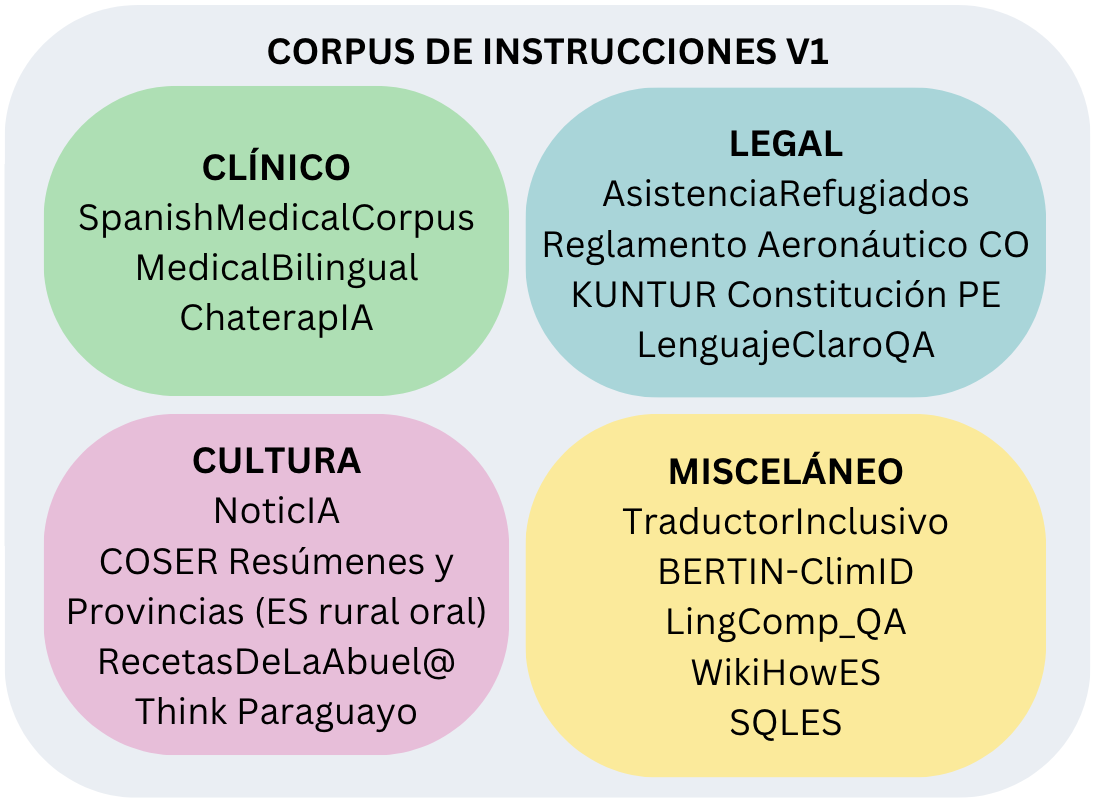}
  \caption{Corpus de instrucciones generados durante el Hackathon \#Somos600M agrupados por dominio.}
  \label{fig:corpus-it-es}
\end{figure}

Destacamos la gran variedad de países representados en los proyectos (e.g., Reglamento Aeronáutico Colombiano, Asistencia a refugiados, Constitución del Perú, recetas típicas), así como el proyecto sobre cultura guaraní.
La mayoría de los equipos se centraron en modelos de texto, excepto un proyecto que se enfocó en las diferentes maneras de hablar en la España rural.
Se generaron gran cantidad de datos de los sectores salud y legal (Figura~\ref{fig:corpus-it-es}).
También destacamos los proyectos de reescritura con lenguaje claro e inclusivo, resumen de noticias clickbait y detección de textos sobre sostenibilidad.

\subsection{Corpus de evaluación}

\begin{figure}[!ht]
  \includegraphics[width=\columnwidth]{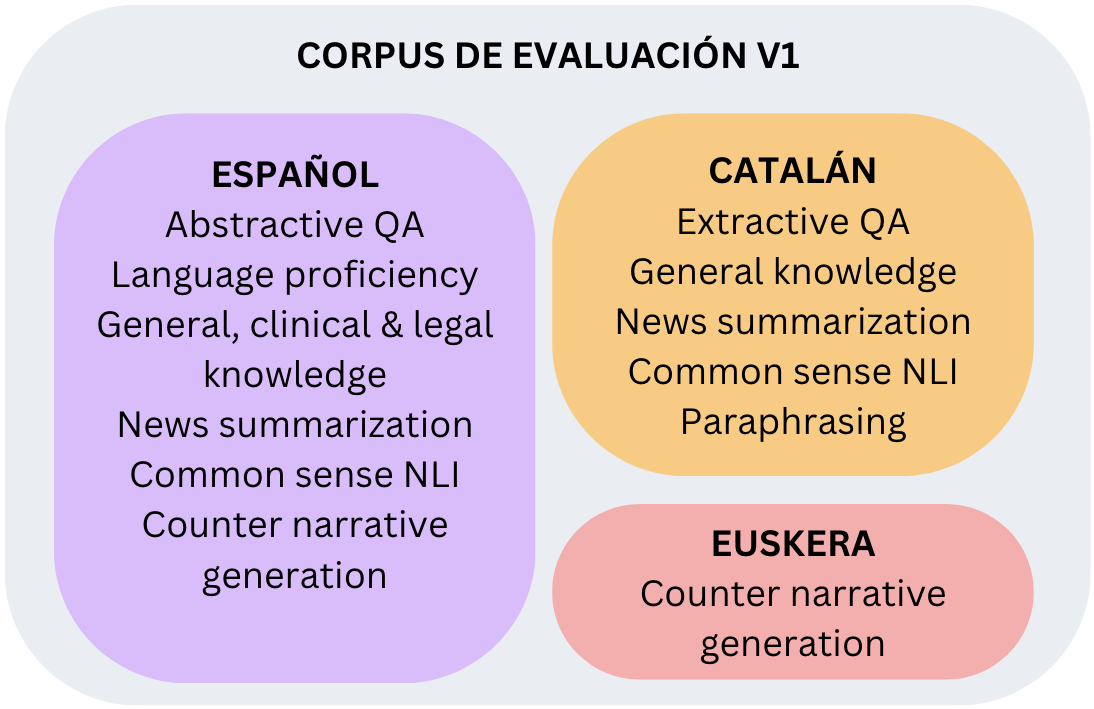}
  \caption{Corpus que constituyen la primera versión de la tabla de clasificación de modelos de lenguaje generativos.}
  \label{fig:corpus-eval-es}
\end{figure}

Con la primera ronda de la campaña de recolección conseguimos la donación de 5 corpus de evaluación anotados manualmente por especialistas y, con la segunda, 14 más. Combinados con las traducciones constituirán la primera tabla de clasificación de LLMs generativos en español (Figura~\ref{fig:corpus-eval-es}).

En la validación de las traducciones de Okapi
participaron un total de 61 personas y se cubrió un 60\% de ARC-C \citep{Clark2018ThinkYH}, 15\% de HellaSwag \citep{zellers2019hellaswag} y 15\% de MMLU \citep{hendrycks2021measuring}. \footnote{hf.co/spaces/somosnlp/BenchmarkAnnotationDashboard}. Con el apoyo de 37 personas se validó el 100\% de las instrucciones de DIBT, convirtiendo al español en el primer idioma en validar al completo su traducción. \footnote{hf.co/spaces/DIBT/PromptTranslationMultilingualDashboard}

\section{Discusión}

Nos llena de satisfacción ver la gran respuesta de la comunidad a nuestra llamada. La creación de 2 millones de instrucciones y la recolección de 22 corpus de evaluación es un gran avance para los LLMs en nuestras lenguas.

Respecto al hackathon, nos alegra ver que el número de bases de datos generados triplica el del año pasado.
Confirmamos la utilidad de las librerías \texttt{distilabel}, \texttt{Argilla} y \texttt{transformers} para el desarrollo de LLMs adaptados con instrucciones sintéticas y revisadas manualmente.

En las campañas de anotación observamos que la mayor parte de la validación fue hecha por un 10\% de las personas. Para esfuerzos similares recomendamos:
1) escribir una guía de anotación clara, habilitar un canal de comunicación y utilizar las dudas para iterar y mejorar la guía,
2) compartir un vídeo de ejemplo
y 3) crear una visualización del progreso de la iniciativa para motivar y dar visibilidad a las personas voluntarias.

\section{Conclusión}
El hackatón, la campaña de recolección y los esfuerzos de anotación nos han permitido crear las primeras versiones del gran corpus de instrucciones y la tabla de clasificación de LLMs generativos.

Vamos a continuar aunando esfuerzos con entidades de LATAM, el Caribe y España para
organizar más hackatones enfocados en temas, variedades o lenguas específicas,
escalar la campaña de recolección para crear un corpus lo más inclusivo posible,
y extender la tabla de clasificación incluyendo evaluaciones de aspectos éticos (sesgos y discurso de odio)
y lingüísticos (e.g. adecuación de la variedad de la lengua generada),
así como otras lenguas cooficiales.

Los recursos generados son abiertos\footnote{huggingface.co/somosnlp}, invitamos a entidades con mayor poder de computación a utilizarlos para entrenar (con nuestro apoyo, si desean) grandes modelos de lenguaje generativos abiertos, de calidad, inclusivos y nativos.

\section*{Agradecimientos}
Agradecemos el esfuerzo de todas las personas participantes en el hackatón, gracias a su trabajo contamos ahora con la primera versión de un gran corpus de instrucciones diverso. Damos gracias a Hugging Face por patrocinar los recursos de computación y a LenguajeNaturalAI, Cálamo \& Cran y SaturdaysAI por patrocinar premios para motivar a los equipos y a LatinX in AI por invitarnos a presentar los proyectos al workshop LatinX in NLP. Gracias también a todas las personas que compartieron su conocimiento con la comunidad en ponencias y talleres.

Respecto a la tabla de clasificación, damos las a Hugging Face y Argilla por co-organizar los esfuerzos de validación de traducciones y a todas las personas voluntarias que participaron en la anotación. También agradecemos las donaciones de bases de datos de evaluación de calidad al Instituto de Ingeniería del Conocimiento (IIC), LenguajeNaturalAI,  Grupo de Internet de Nueva Generación (GING) de la Universidad Politécnica de Madrid (UPM), Centro Vasco de Tecnología de la Lengua (HiTZ) y Barcelona Supercomputing Center (BSC). 

Gracias a todas las personas que revisaron este artículo, especialmente a Diana Galván, Flor Plaza, Abi Oppenheim, Pedro Reviriego, Javier Conde y Javier Aula-Blasco.

Para finalizar, damos las gracias de corazón a todas las personas que voluntariamente ofrecen su tiempo para apoyar nuestra misión de democratizar el PLN para la comunidad hispanohablante.

\appendix

\section{Recursos de español en el Hub de Hugging Face}
\label{sec:appendix-hf-hub-es}

\begin{figure}[!ht]
  \includegraphics[width=\columnwidth]{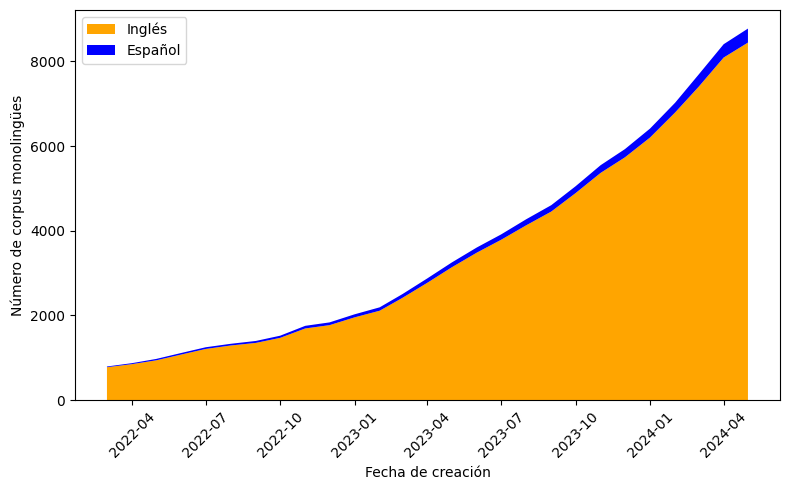}
  \caption{Evolución del número de corpus monolingües en inglés (naranja) y español (azul) en el Hub de Hugging Face hasta el 13 de mayo de 2024.}
  \label{fig:huggingface-hub-es}
\end{figure}

Aunque el número de recursos abiertos de PLN en español en el Hub de Hugging Face esté aumentando, la brecha entre el español y el inglés sigue siendo inmensa (Figura~\ref{fig:huggingface-hub-es}).

El código utilizado para visualizar esta brecha está disponible en el Hub de Hugging Face\footnote{hf.co/spaces/mariagrandury/language-gap-in-hf-hub}.

\section{Participantes del hackatón}
\label{sec:appendix-hackathon-es}

651 personas de 29 países ((Figura~\ref{fig:map}) se registraron para el hackatón, de las cuales 222 estaban interesadas únicamente en asistir a las charlas. Cabe destacar que como mínimo\footnote{Las preguntas no son de obligatoria respuesta.} el 14\% de las personas ya habían participado en alguna de las dos ediciones anteriores, casi el 50\% se conectaba desde LATAM (Tabla~\ref{tab:location}) y menos del 40\% se auto-identifica con un género diferente al masculino (Tabla~\ref{tab:gender}), nos gustaría incrementar estos tres números en próximas ediciones. Respecto al nivel previo y la ocupación, más del 40\% de las personas afirmaron tener un nivel de PLN fundamental antes de comenzar el hackatón (Tabla~\ref{tab:level}) y la mayor parte son desarrolladores/as, ingenieros/as o científicos/as de datos (Figura~\ref{fig:wordcloud}).

\begin{figure}[!ht]
    \includegraphics[width=\columnwidth]{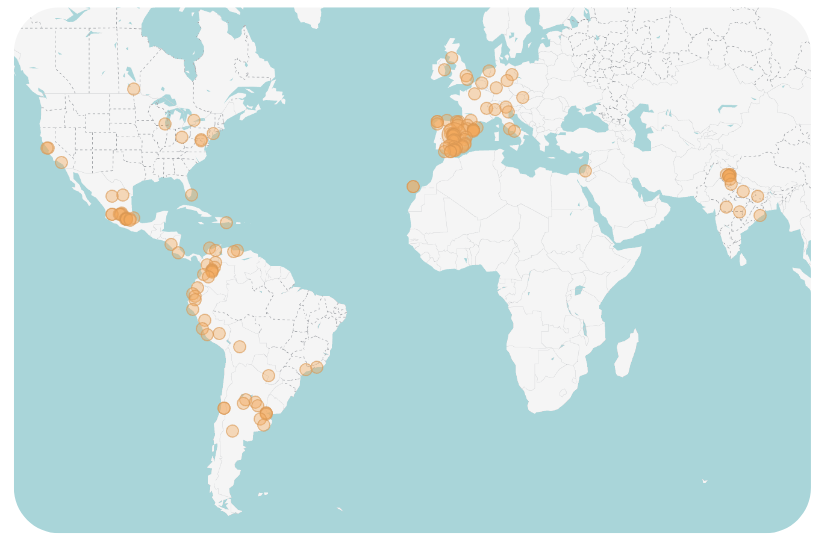}
  \caption{Localización de las personas participantes en el Hackathon \#Somos600M.}
  \label{fig:map-es}
\end{figure}

\begin{table}[!ht]
   \centering
  \begin{tabular}{ccc}
    \hline
    \textbf{ LATAM } & \textbf{ España } & \textbf{ Otros }  \\
    \hline
    46\% & 32\% & 22\% \\
    \hline
  \end{tabular}
  \caption{Localización de las personas participantes en el Hackathon \#Somos600M, puede no coincidir con la nacionalidad.}
  \label{tab:location-es}
\end{table}

\begin{table}[!ht]
   \centering
  \begin{tabular}{cccc}
    \hline
    \textbf{ Femenino } & \textbf{ Masculino } & \textbf{ NB } & \textbf{ SR }  \\
    \hline
    22\% & 60\% & 1\% & 17\% \\
    \hline
  \end{tabular}
  \caption{Género con el que se auto-identifican las personas participantes en el Hackathon \#Somos600M: "femenino", "masculino", "NB" (no binario) o "SR" (sin respuesta).}
  \label{tab:gender-es}
\end{table}

\begin{table}[!ht]
   \centering
  \begin{tabular}{cccc}
    \hline
    \textbf{ Fund. } & \textbf{ Intermedio } & \textbf{ Avanzado } & \textbf{ SR }  \\
    \hline
    40\% & 31\% & 12\% & 17\% \\
    \hline
  \end{tabular}
  \caption{Nivel auto-asignado de las personas participantes en el Hackathon \#Somos600M: "fundamental", "intermedio", "avanzado", "SR" (sin respuesta)"}
  \label{tab:level-es}
\end{table}

\begin{figure}[!ht]
    \includegraphics[width=\columnwidth]{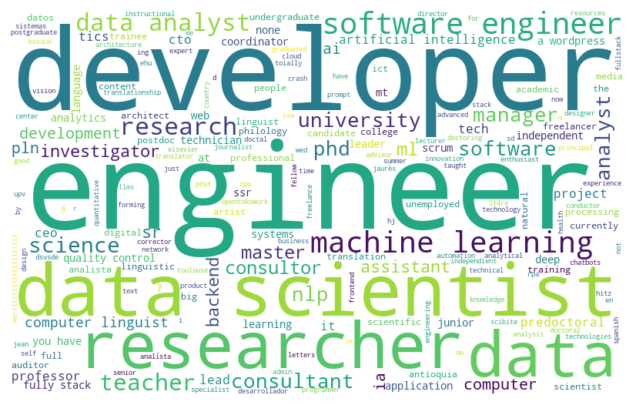}
  \caption{Nube de palabras representando las ocupaciones de las personas que asistieron al Hackathon \#Somos600M, previamente traducidas al inglés para combinar términos equivalentes.}
  \label{fig:wordcloud-es}
\end{figure}

\section{Guías y recursos}
\label{sec:appendix-hackathon-resources-es}

Las guías del hackatón\footnote{somosnlp.org/hackathon/bases} y de la campaña de anotación\footnote{somosnlp.org/blog/validacion-benchmarks-argilla} están disponibles en la página de SomosNLP.

Dado que el hackatón estaba abierto a todo el mundo y se esperaban participantes de todos los niveles, se organizaron sesiones de mentoría, talleres aplicados y charlas avanzadas. Las grabaciones de todas las ponencias están disponibles en YouTube\footnote{youtube.com/watch?v=JzpvHRrqtSU\&list=PLTA-KAy8nxaASMwEUWkkTfMaDxWBxn-8J} y los cuadernos correspondientes, en GitHub\footnote{github.com/somosnlp/recursos/tree/main/hackathon\_2024}.

\section{Detalle de los corpus del hackatón}
\label{sec:appendix-corpus-it-es}

Enumeramos los corpus creados durante el Hackathon \#Somos600M, detallados en la Tabla~\ref{tab:instructions-es}:
\textbf{AsistenciaRefugiados}: Asistente legal para personas en situación de refugio o asilo político \citep{somosnlp2024asistenciarefugiados},
\textbf{BERTIN-ClimID}: BERTIN-Base Climate-related text Identification \citep{huerta2024climid},
\textbf{ChaterapIA}: Conversaciones terapéuticas \citep{juliofc2024convterapeuticas},
\textbf{COSER Provincias}: Indentificación de provincias a partir de entrevistas orales de la España rural \citep{adsuar2024coserprovincias},
\textbf{COSER Resúmenes}: Resumen de entrevistas orales de la España rural \citep{adsuar2024coserresumenes},
\textbf{KUNTUR}: Asistencia legal en textos jurídicos de Perú \citep{quispe2024kuntur},
\textbf{LenguajeClaroQA}: Simplificación de lenguaje administrativo \citep{delafuente2024lenguajeclaro},
\textbf{LingComp\_QA}: Corpus educativo de lingüística computacional en español \citep{LingComp_QA},
\textbf{NoticIA}: Resumen de Noticias Clickbait \citep{garcíaferrero2024noticia},
\textbf{RAC}: Reglamento Aeronáutico Colombiano \citep{sepulveda2024rac},
\textbf{RecetasDeLaAbuel@}: Corpus de recetas de países hispanoamericanos \citep{recetasdelaabuela2024},
\textbf{SMC}: Spanish Medical Corpus \citep{lopez2024spanishmedicallm},
\textbf{SQLES}: Interactúa con una base de datos SQL en español \citep{sql2024rangel},
\textbf{ThinkParaguayo}: Conoce la cultura guaraní \citep{paiva2024guarani},
\textbf{TraductorInclusivo}: Reescritura de textos en español utilizando lenguaje inclusivo \citep{GAMIJ2024es-inclusive-language},
y
\textbf{WikiHowES} \citep{quispe2024wikihowes}.

\begin{table*}[p]
   \centering
  \begin{tabular}{lcccc}
    \hline
    \textbf{Corpus} & \textbf{Nº ejemplos} & \textbf{MB} & \textbf{Dominio} & \textbf{País(es)} \\
    \hline
    AsistenciaRefugiados & 10707 & 20.7 & Legal & ES, MX, VE + \\
    BERTIN-ClimID & 3680 & 1.63 & Sostenibilidad & PE, ES \\
    ChaterapIA & 1000 & 2.30 & Psicología & ES \\
    COSER Provincias & 1150 & 0.22 & Cultura rural & ES (oral) \\
    COSER Resúmenes & 230 & 1.08 & Cultura rural & ES (oral) \\
    KUNTUR & 2075 & 0.73 & Legal & PE \\
    LenguajeClaroQA & 4094 & 1.72 & Legal admin. & ES \\
    LingComp\_QA & 1004 & 0.35 & Lingüística Comp. & ES \\
    MedicalBilingual & 8138 & 12.8 & Clínico & Mix \\
    NoticIA & 850 & 3.41 & Prensa & ES \\
    RAC & 24478 & 1.84 & Legal & CO \\
    RecetasDeLaAbuel@ & 20221 & 42.4 & Gastronomía & ES, MX, PE, AR+ \\
    SpanishMedicalCorpus & 2136490 & 48.5 & Clínico & ES, CL \\
    SQLES & 81 & 0.40 & Programación & - \\
    Think Paraguayo & 1498 & 0.19 & Cultura guaraní & PY \\
    TraductorInclusivo & 4196 & 0.40 & Misceláneo & ES, AR, MX, CL, CR + \\
    WikiHowES & 113160 & 186 & Misceláneo & Mix \\
    \hline
    Total & 2,333,052 & 324.67 & - & - \\
    \hline
  \end{tabular}
  \caption{Corpus de instrucciones creados por los equipos participantes en el Hackathon \#Somos600M, disponibles en huggingface.co/somosnlp. Se excluyen las versiones de los corpus adaptadas a formatos para el entrenamiento de un modelo específico (e.g. Gemma). Los países están representados por su correspondiente código ISO 3166-1 alfa-2, el signo "+" indica que hay más países representados en el corpus en menor proporción.}
  \label{tab:instructions-es}
 
\end{table*}

\section{Detalle de los corpus de evaluación}
\label{sec:appendix-corpus-eval-es}

A continuación se enumeran los corpus de evaluación incluidos en la primera versión de la tabla de clasificación de modelos generativos.
\textbf{AQuAS}: Abstractive Question-Answering in Spanish \citep{iic2024aquas} y \textbf{RagQuAS}: Retrieval-Augmented-Generation and Question-Answering in Spanish \citep{iic2024ragquas}, donados por el Instituto de Ingeniería del Conocimiento (IIC) de la Universidad Autónoma de Madrid (UAM);
\textbf{SpaLawEx}: Exámenes de acceso a la abogacía \citep{lenguajenaturalai2024spalawex}, \textbf{MedicalExpertES}: Diagnóstico y tratamiento de casos clínicos en español \citep{lenguajenaturalai2024medicalexpertes} y \textbf{HumorQA}: Clasificación de bromas de humor blanco \citep{lenguajenaturalai2024humorqa}, donados por LenguajeNaturalAI;
\textbf{TELEIA}: Test de Español como Lengua Extranjera para Inteligencia Artificial, donado por el Grupo de Internet de Nueva Generación (GING) de la Universidad Politécnica de Madrid (UPM);
\textbf{Meta4XNLI}: A Crosslingual Parallel Corpus for Metaphor Detection and Interpretation \citep{sanchezbayona2024meta4xnli}, \textbf{NoticIA}: Resumen de Noticias Clickbait \citep{garcíaferrero2024noticia}, \textbf{MedExpQA}: Multilingual Benchmarking of Large Language Models for Medical Question Answering
\citep{alonso2024medexpqa}, \textbf{CasiMedicos-SQUAD}: Explanatory Argument Extraction of Correct Answers in Resident Medical Exams \citep{goenaga2023explanatory} and \textbf{CONAN-EUS}: Basque and Spanish Parallel Counter Narratives Dataset\citep{bengoetxea-et-al-2024} donados por el Centro Vasco de Tecnología de la Lengua (HiTZ);
\textbf{CatalanQA}: Extractive QA in Catalan, \textbf{TE\_ca}: Textual Entailment in Catalan, \textbf{caBREU}: Article Summarization in Catalan, \textbf{XNLI\_ca}: Cross-lingual Natural Language Inference in Catalan, \textbf{COPA\_ca}: Choice Of Plausible Alternatives in Catalan, \textbf{PAWS\_ca}: Paraphrase Adversaries from Word Scrambling in Catalan, \textbf{XQUAD\_ca}: Cross-lingual Question Answering Dataset in Catalan and \textbf{WNLI\_ca}: Winograd NLI dataset, creados o traducidos profesionalmente con financiación de los proyectos AINA e ILENIA y donados por el Barcelona Supercomputing Center (BSC) \citep{gonzalez-aguirre-etal-2024-infraestructure}.

\begin{table*}[!ht]
  \centering
  \begin{tabular}{lccc}
    \hline
    \textbf{ Corpus } & \textbf{ Lengua } & \textbf{ Dominio } & \textbf{ Tarea } \\
    \hline
    AQuAS & ES & Misceláneo & Extracción de información \\ 
    RagQuAS & ES & Misceláneo & RAG y extracción de información \\ 
    HellaSwag\_es & ES & Misceláneo & Commonsense NLI \\ 
    MMLU\_es & ES & Misceláneo & Preguntas de opción múltiple \\
    TELEIA & ES & Dominio idioma & Preguntas de opción múltiple \\
    Meta4XNLI & ES & Dominio idioma & NLI \\ 
    HumorQA & ES & Dominio idioma & Clasificación \\
    ARC-C\_es & ES & Ciencia & Preguntas de opción múltiple \\
    NoticIA & ES & Prensa & Resumen de texto \\
    SpaLawEx & ES & Legal & Preguntas de opción múltiple \\
    MedicalExpertES & ES & Clínico & Preguntas de respuesta abierta \\
    MedExpQA & ES & Clínico & Preguntas de opción múltiple \\
    CasiMedicos-SQUAD & ES & Clínico & Extracción de información \\
    CONAN-EUS & ES, EU & Discurso de odio & Generación de contranarrativas \\
    CatalanQA & CA & Misceláneo & Extracción de información \\ 
    TE\_ca & CA & Misceláneo & NLI \\ 
    XNLI\_ca & CA & Misceláneo & NLI \\ 
    WNLI\_ca & CA & Misceláneo & NLI \\ 
    COPA\_ca & CA & Misceláneo & Razonamiento lógico \\ 
    PAWS\_ca & CA & Misceláneo & Paraphrasing \\
    XQUAD\_ca & CA & Misceláneo & Extracción de información \\ 
    caBREU & CA & Prensa & Resumen \\
    \hline
  \end{tabular}
  \caption{Corpus que constituyen la primera versión de la tabla de clasificación de modelos de lenguaje generativos que incluye tareas en español (ES), catalán (CA) y euskera (EU) y evalúa la capacidad de extracción de información, cultura general, conocimiento en los dominios legal y clínico, razonamiento lógico y dominio del idioma. 
  }
  \label{tab:corpus-eval-es}
\end{table*}

\end{document}